# *Predicting Gene-Disease Associations with Knowledge Graph Embeddings over Multiple Ontologies*

*Susana Nunes, Rita T. Sousa, Catia Pesquita*

*LASIGE, Faculdade de Ciências, Universidade de Lisboa*

*ABSTRACT

Ontology-based approaches for predicting gene-disease associations include the more classical semantic similarity methods and more recently knowledge graph embeddings. While semantic similarity is typically restricted to hierarchical relations within the ontology, knowledge graph embeddings consider their full breadth. However, embeddings are produced over a single graph and complex tasks such as gene-disease association may require additional ontologies. We investigate the impact of employing richer semantic representations that are based on more than one ontology, able to represent both genes and diseases and consider multiple kinds of relations within the ontologies. Our experiments demonstrate the value of employing knowledge graph embeddings based on random-walks and highlight the need for a closer integration of different ontologies.

## 1 INTRODUCTION

Genomic studies and high-throughput experiments often produce large lists of candidate genes, of which only a small portion is truly relevant to the disease, phenotype, or biological process of interest. Deciphering the link between genes and diseases is one of the most demanding tasks in the biomedical field of expertise. Complex diseases present a highly heterogeneous genotype, which hinders biological marker identification. Computational approaches for predicting gene-disease associations can greatly accelerate this process but must tackle several challenges, such as obtaining associations on a large scale while at the same time not compromising on the quality of the associations. In this context, recent developments that explore the scientific knowledge described in ontologies have achieved good results [1].

State-of-the-art approaches in this area are typically based on Semantic Similarity Measures (SSM). Zeng et al. (2017) used phenotype–phenotype similarity to prioritize novel gene–phenotype associations [2]. Asif et al. (2018) show that machine learning classifiers trained on gene functional similarities, using Gene Ontology, can improve the identification of genes involved in complex diseases [3]. However, classical SSM only consider hierarchical relations when measuring the similarity between entities described with ontology concepts [4].

When biomedical entities are annotated with ontology concepts, we can consider them to constitute a Knowledge Graph (KG). KG embeddings (KGEs) allow us to represent each entity with a vector that approximates the similarity properties of the graph [5,6]. KGEs support in principal more powerful representations that consider multiple types of relations. Recent advances in this are such as Onto2Vec [7] and OPA2Vec [8] generate vector representations of biological entities in ontologies by combining formal ontology axioms and annotation axioms from the ontology metadata. KGE work over a single integrated graph, however, being limited to a single ontology may be insufficient in multi-domain applications such as gene-disease associations.

In this work, we investigate the impact of employing richer semantic representations in gene-disease association prediction. The semantic representations we propose are enriched because they are based on more than one ontology, they are able to represent both genes and diseases and they consider multiple kinds of relations within the ontologies. We explore the Human Phenotype Ontology (HP) [9] and the Gene Ontology (GO) [10] to represent gene and disease features in a shared semantic space that covers both gene function and phenotypes. HP provides a standardized vocabulary of phenotypic abnormalities encountered in human disease and GO describes the function of genes and gene products. We hypothesize that using more than one ontology can improve gene-disease association prediction and that a richer integration can have a positive impact. In particular, we consider Logical Definitions (LD) that associate entities from different ontologies with complex semantic relations can be explored to bridge domains and contextualize relations between different entities, such as a gene and a disease. For example, the HP term for "Hearing impairment" (HP:0000365) is equivalent to a restriction that involves the GO term "Sensory perception of sound" (GO:0007605). We also investigate multiple KGE methods to represent both genes and diseases and how these vectors can be combined.

*Correspondence: snunes@lagise.di.fc.ul.pt

## 2 METHODS

The first step in our approach is to integrate the different ontologies and annotation data to build the KGs. Then we produce embeddings that represent the gene and the disease according to their annotations in different KGs. These embeddings are combined using different vector operators producing a representation of genes and diseases in what is effectively a shared semantic space. Finally, supervised learning algorithms are trained over the combined embeddings to predict gene-disease associations. This approach is evaluated against non-ML approaches based on classical SSM and KGE similarity[1].

### 2.1 Dataset and Annotations

First, we obtained all curated gene-disease associations from DisGeNET - one of the largest available collections of genes and variants involved in human disease [11]. To avoid bias regarding annotations, we proceeded to filter these pairs considering only relations where the source did not rely on the databases Uniprot [12], OMIM [13], or Orphanet [14]. The 84038 resulting pairs were then filtered according to three criteria: i) the genes must have correspondence with a protein from Uniprot that is annotated with GO terms; ii) the genes must be annotated with HP terms; iii) the diseases must be annotated with HP terms. A total of 2716 genes, 1807 diseases, and 8189 disease-genes relations remained. Considering that negative samples are not included in DisGeNET, we employed random sampling to create negative examples composed of the genes and diseases present in the positive examples, but without known associations between them, building a balanced dataset.

GO annotations were downloaded from Gene Ontology Annotation[2] (GOA) database for the human species[3][15]. HP annotations were downloaded from the HP[4] database, providing links between genes or diseases to HP terms [9].

### 2.2 Knowledge Graphs

The KGs used in this work are divided into three possible types:
(1) **HP:** HP (without LDs) and HP annotations both for genes and diseases.
(2) **HP + GO:** HP without LDs, HP annotations (for genes and diseases), GO and GO annotations (for genes). HP and GO are integrated through a common virtual root.
(3) **HP + LD + GO:** HP+GO with added LDs. To simplify graph embeddings approaches, LDs were transformed into simple equivalence statements between the HP class and the GO class contained in the LD.

### 2.3 Baseline

The baseline aims to establish the performance of methods that use a single ontology and classical SSM. We used the HP KG since it annotates both genes and diseases and measured semantic similarity for all gene-disease pairs using six different SSM: $BMA_{ICSeco}$ [16], $BMA_{ICResnik}$ [17], $SimGIC_{ICSeco}$ [16,18], $SimGIC_{ICResnik}$ [17,18], $Max_{ICSeco}$ [16] and $MAX_{ICResnik}$ [17].

The association prediction is expressed as a classification problem where a SS score for a gene-disease pair exceeding a certain threshold indicates a positive association. For each measure, a SS threshold was chosen after evaluating the weighted average of F-measures (for positive and negative predictions) at different thresholds and selecting the maximum (values in the range from 0 to 1 with a step of 0.01). The baselines represent the best possible result obtained with SSM and HP.

### 2.4 Knowledge Graph Embeddings

We generated KGEs with 200 features and used five different methods that cover different approaches for KGE: translational distance (TransE [19]), semantic matching (DistMult [20]) and random walk-based (RDF2Vec [21], OPA2Vec [8], OWL2Vec* [22]).

### 2.5 Vector Representation and Prediction

Each gene-disease pair corresponds to two vectors, *g* and *d*, associated with a gene and a disease, respectively. We defined an operator over the corresponding vectors in order to generate a representation *r(g,d)*. Several choices for the operator were considered: Concatenation, Average, Hadamard, Weighted-L1, Weighted-L2. The resulting vectors were then the input to four different ML algorithms: Random Forest (RF) [23], eXtreme Gradient Boosting (XGB) [24], Naïve Bayes [25] and Multi-Layer Perceptron (MLP) [26]. Grid search was employed to obtain optimal parameters for RF, XGB, and MLP.

We also measured the cosine similarity (CS) between the vectors carrying out the same approach used in the baseline with a SS threshold.

### 2.6 Performance measure

The classification performance was evaluated using the weighted average of F-measures (WAF). In addition to the WAF, we evaluated the performance with AUC (area under of the receiver operating characteristic (ROC) curve) scores. In each experiment, we performed a stratified 70% training and 30% testing split method, being that the same split was used throughout all experiments, including in the baseline.

---

[1] All data and code available at: https://github.com/liseda-lab/KGE_Predictions_GD
[2] GOA version date 11 August of 2020
[3] GO version date 8 December 2020
[4] HP version date 12 October 2020

## 3 RESULTS

### 3.1 Baseline Performance

The performance of the six SSMs using the HP KG is shown in **Table 1**. BMA$_{ICResnik}$ and MAX$_{ICResnik}$ achieve the best performance for a single ontology, both with a 0.693 WAF score. The measure chosen as the final baseline was best performing in terms of AUC score: BMA$_{ICResnik}$ with 0.733 against MAX$_{ICResnik}$ with 0.731.

**Table 1.** WAFs scores for optimal SSM performance with HP ontology.

| BMA | BMA | SimGIC | SimGIC | MAX | MAX |
|---|---|---|---|---|---|
| ICSeco | ICResnik | ICSeco | ICResnik | ICSeco | ICResnik |
| 0.689 | **0.693** | 0.655 | 0.655 | 0.639 | **0.693** |

In bold, the best results of WAF with SSM.

### 3.2 Rich Semantic Representations Performance

There are three important aspects in elucidating the performance impact of the different factors in richer semantic representations: (1) how can we combine the gene and disease vectors; (2) which KGE methods are more suitable for this task; (3) what is the impact of considering more than one ontology.

**Figure 1** compares the 5 vector operations illustrated with the three best performing KGE methods (RDF2Vec, OPA2Vec and DistMult) coupled with Random Forest (one of the best performing ML methods). The Hadamard operator outperforms other operators with differences being more evident when using OPA2Vec.

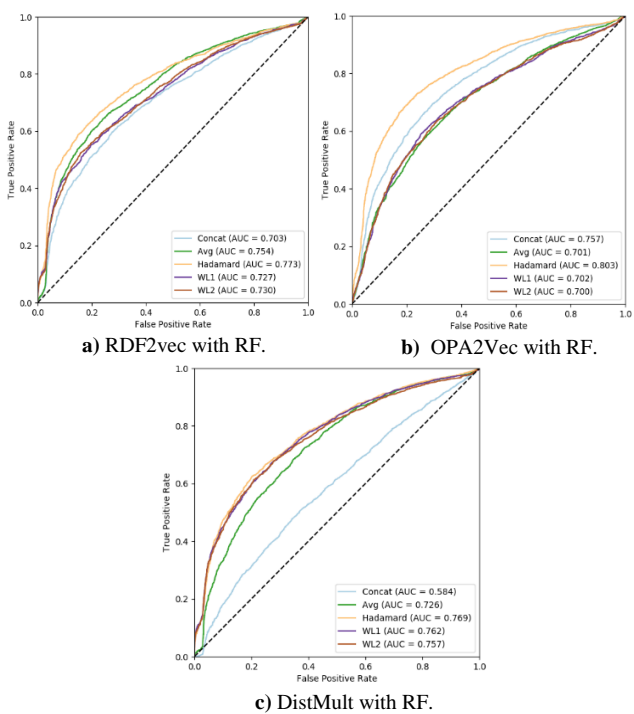

a) RDF2vec with RF.  b) OPA2Vec with RF.

c) DistMult with RF.

**Figure 1.** ROC curves and AUC values obtained for different vector operators with RF classifier for the HP+GO KG.

**Table 2** presents the performance obtained for the three KGs considered with the best performing KGE and ML approaches using the Hadamard operator as well as the cosine similarity-based approach. When using CS, the differences between the KGE-based approaches and the best baseline SSMs are reduced. However, when using ML approaches, RDF2Vec, OPA2Vec and DistMult achieve better results than the SSM baselines by up to 11%. On one hand, reducing the representation to a similarity score may be too limiting, on the other KGE methods afford richer representations by both employing the full breadth of relations in the KG and providing multi-dimensional representations, something that SSMs do not. Moreover, TransE achieved poorer results because they are less well suited to capture longer distance relations than random-walk and semantic matching methods, something that is perhaps more relevant in the context of ontologies (where hierarchical relations between ontology classes are the backbone) than in the KG context (where links between entities are the focus).

**Table 2.** WAFs for each competing combination of KGEs and ML algorithms for different KGs and Hadamard operator.

|  |  | HP | HP + GO | HP + GO + LD |
|---|---|---|---|---|
| **RDF2Vec** | CS | 0.682 | 0.685 | 0.681 |
|  | RF | 0.717 | 0.723 | 0.716 |
|  | XGB | 0.709 | 0.711 | 0.715 |
| **OPA2Vec** | CS | 0.694 | 0.667 | 0.683 |
|  | RF | 0.736 | 0.741 | **0.745** |
|  | XGB | 0.739 | 0.739 | 0.742 |
| **DistMult** | CS | 0.706 | 0.689 | 0.702 |
|  | RF | 0.715 | 0.701 | 0.706 |
|  | XGB | 0.717 | 0.705 | 0.708 |

In bold, the best WAF score achieved within the various combinations.

Furthermore, OPA2Vec achieves the best results and is also able to take advantage of the integration of ontologies with a small improvement in performance. The best combination of methods achieves 0.745 WAF by using HP+GO+LD KG with OPA2Vec and the RF Classifier. OPA2Vec generates a corpus with a set of asserted and inferred axioms from an ontology, the set of annotation axioms involving labels, descriptions, synonyms and creators, and the gene and disease phenotype annotations. By including all these features, this is a probable cause to present itself as a more stable and successful algorithm over the others.

## 4 CONCLUSION

Associating genes with diseases is an important area of research because it is a crucial challenge in human health with applications to understand disease etiology and develop new techniques for prevention, diagnosis, and therapy.

In this study, we have proposed a novel approach to predict gene-disease associations using rich semantic representations based on KGEs over multiple ontologies. We investigated the impact of different approaches to build a shared rich semantic representation for genes and diseases, as well as multiple KGE methods and ML algorithms.

We have shown that employing KGEs combined with the Hadamard operator outperforms classical SSM achieving an AUC score of 0.821. However, our experiments reveal that differences between using a single ontology or combining two ontologies (HP and GO) are comparatively small regardless of a richer integration using LDs. We hypothesize that the information provided by the LDs does not provide substantial additional information comparing with what is already present in HP. This may partially be explained by the existence of only 351 LDs that link HP and GO. Our next steps will explore ontology matching techniques to create additional logical definitions and links between ontologies [27], which will also allow us to expand the ontologies used to those that do not contain LDs between them.

## ACKNOWLEDGEMENTS

This work was supported by FCT through LASIGE Research Unit (ref. UIDB/00408/2020 and ref. UIDP/00408/2020. It was also partially supported by the KATY project which has received funding from the European Union's Horizon 2020 research and innovation program under grant agreement No 101017453.